\documentclass[10pt, conference]{ieeeconf}

\usepackage{microtype}
\usepackage{graphicx}
\usepackage{subfigure}
\usepackage{booktabs} 
\usepackage{tabularx}
\usepackage{multirow}  
\usepackage{multicol}

\usepackage{amsmath} 

\usepackage{algorithm}
\usepackage{algorithmic}

\usepackage{amsfonts}
\usepackage{caption}
\usepackage{xcolor}
\usepackage{float}

\author{
Pierrick Pochelu \\
University of Lille\\
Pau, France\\
pierrick.pochelu@gmail.com \\
\and
Serge G. Petiton \\
University of Lille, CNRS \\ 
Lille, France\\
serge.petiton@univ-lille.fr \\
\and
Bruno Conche \\
TotalEnergies SE \\
Pau, France\\
bruno.conche@totalenergies.com \\
}

\title{\Large{Distributed Ensembles of Reinforcement Learning Agents for Electricity Control}}

\begin{document}
\maketitle

\thispagestyle{empty}
\pagestyle{empty}
\vspace{-2cm}
\begin{abstract}


Deep Reinforcement Learning (or just “RL”) is
gaining popularity for industrial and research applications.
However, it still suffers from some key limits slowing down
its widespread adoption. Its performance is sensitive to initial
conditions and non-determinism. To unlock those challenges,
we propose a procedure for building ensembles of RL agents to efficiently build better local decisions toward long-term
cumulated rewards. For the first time, hundreds of experiments
have been done to compare different ensemble constructions
procedures in 2 electricity control environments. We discovered
an ensemble of 4 agents improves accumulated rewards by 46\%, improves reproducibility by a factor of 3.6, and can naturally
and efficiently train and predict in parallel on GPUs and
CPUs.

\end{abstract}

\section{Introduction}
Electricity control is a challenging subject due to many aspects: intermittent nature of renewable energy, variations in demand, low storage abilities, \cite{rlpym} \cite{rlanm:2021} significant room for improvement for running electric engines \cite{rlgem}. Deep Reinforcement Learning has shown great success in scaling up model-free reinforcement learning algorithms to the challenging Markov Decision Processes \cite{mnih:2015} \cite{silver:2016} and is a promising method to solve issues of electricity control \cite{rlapplications:2021}. However, reinforcement learning is sensitive to random effects and hyperparameters tuning \cite{rlmatter} making them difficult to use in practice.

To alleviate this, we analyze and propose an ensemble of deep reinforcement learning agent procedures and discuss its performance on both sides: quality of automatic decisions and the computing cost at training time and inference time. We discovered that an ensemble of off-policy RL agents overpasses the simple baseline consisting to select the best single agent in it. And more, ensembles are generally worth the computing cost on recent hardware without systematically requiring GPU investment.

Ensemble learning is now well-known for producing qualitative predictions by using the instability of the same learners at different run times \cite{useoverfit:1995}. In the context of RL, it has the potential to improve the robustness and quality of reinforcement learning trajectories by taking a better sequence of local better decisions. 

In the last few years, some ensembles of reinforcement learning have been proposed and demonstrate good abilities \cite{ensrldqn} \cite{rlmetrpo:2018} to increase the diversity of samples gathered to improve the stability and to accelerate the training process compared to one single agent. However, there are still open research questions to be investigated: which procedure is better to build them? Does ensemble worth the computing cost compared to selecting the best model in it?

In this paper, we aim to answer them. We will not limit ourselves to comparing the training dynamic of two ensembles or comparing one agent to one single agent. We instead sweep hundreds of hyperparameters to fairly compare ensembles by splitting training, validation, and testing phases to reflect the need to measure how well they generalize. For the first time, multiple ensemble procedures are compared: homogeneous \cite{useoverfit:1995} (same hyperparameters but different random seeds) and heterogeneous \cite{heteroinput:1999} \cite{creatediversity:2005} (different hyperparameters different random seeds) aggregation procedures. And different combination rules, averaging\cite{useoverfit:1995}, weighted averaging \cite{wavg:1994}, and select only the best base agent (e.i., no ensemble). Then, we evaluate the computing cost of the building phase and the inference phase running on modern computing nodes.


This paper first demonstrates experimental evidence that homogeneous ensembles with averaging as a combination rule are more performant and stabler than one individual RL agent and other ensemble procedures. Second, we perform extended experiments by increasing the number of agents and conclude that it significantly improves the stabilization of the cumulative reward without widely increasing the computation time of modern GPUs and CPUs. Finally, due to the simplicity of the proposed procedure and the stabilization effects, our experiments are easily reproducible. 

We experiment with diverse ensembles of RL agent constructions on two industrial applications linked to electricity control: Pymgrid \cite{rlpym} and GEM \cite{rlgem}. Both are available with a continuous and discrete action space. We use off-policy RL algorithms: DQN for discrete environments \cite{rlq} \cite{rldueling} and DDPG \cite{rlddpg} for continuous ones. However, it is still unsure if ensembling would be beneficial to other RL algorithms. Those results may not generalize well in some RL frameworks where RL agents lack exploration, the interpolation of too many specialized agents may drive the ensemble in some trajectories where agents have low experiences. And more, in this work, we interpolate predictions as combination rules, which makes sense in the electricity control problems but application to other environments may require a domain-specific combination rule.

Our paper follows this structure: In section II we introduce the workflow we use to perform ensemble benchmarks. In sections III and IV we perform multiple benchmarks to assess different ensemble construction procedures in terms of test scores and their stability.

\section{The proposed ensemble of RL agents}

To solve a given control problem, Random Search to generate multiple Reinforcement Learning hyperparameters given an RL framework is a simple and robust method \cite{random:2020}. In this paper, hyperparameters will not encode only one agent but the ensemble of agents.

Offline RL agents have two different policies depending on whether they are in the training or evaluation phase.  The agent is trained with an explorative-exploitative behavior to learn to control the environment. Once the agent is trained (i.e., inference phase), the agent acts with its best guess in the environment (exploitation only). Multiple training and evaluation are needed to measure the performance and the robustness of one given hyperparameter.

The procedure we follow is described as follows: 1) Hyperparameters are sampled with Random Search to build hundreds of ensembles. There are multiple variants we assess: either homogeneous ensembles or heterogeneous ensembles, and different combination rules 2) In each ensemble, all base agents are trained independently, and evaluated individually to produce an individual validation score. 3) Ensembles may be calibrated according to the individual validation score to calibrate the combination rule. Then, the validation score of the ensemble is computed. 4) The ensemble performing the best in step 3 is taken and deployed in production. We simulate the final performance in production by testing it the last time. It is the test score of the ensemble.

\subsection{Training, validation, and testing}

Many reinforcement learning projects applied to some tasks such as video games or playboards are only interested in maximizing rewards during the training loop and comparing RL approaches based on the training dynamics. We rather want to train our agents on a given runtime and measure how well they generalize. The method must be robust to the initial conditions such as different random seeds. We also want to avoid overfitting the training transitions stored in the replay buffer and avoid overfitting the time series data associated with the environment (Pymgrid contains some time series but not GEM). This is why we propose that the base agent training loops are independent with no sharing of the experience memory. No sharing of collected data samples allows them to build their own experiences and statistically maximize their diversity.

It is well known that most supervised machine learning projects have a training phase and two evaluations phase: the validation and the test which are different subsets of data samples from the training phase. We use the same terminology to make good quality decisions in the context of Reinforcement Learning.
\begin{itemize}
    \item The validation score allows making decisions on the neural network, like selecting the best agent given this score, calibration of the ensemble, early stopping \cite{early:1998} (not assessed here) ...
    \item The test score allows simulating the true score when it will be deployed in production. It may provide different values from the validation for multiple reasons: different random seeds, different number of states to evaluate (``different horizon''), and different time-series data (if any).
\end{itemize}

We name ``R'' the evaluation score (``test score'' when it is not precised). This is the cumulative rewards during the evaluation phase. $R=\sum_{t=1}^{h}r_{t}$ for each time step $t$ returning the reward $r_{t}$ until the horizon $h$ is reached.

\subsection{Ensemble construction}

Homogeneous ensembles are built with $n$ hyperparameters, each agent shares the same $n$ hyperparameters but different random seeds. In heterogeneous ensembles of $m$ agents, the ensemble has $m*n$ hyperparameters.

Table~\ref{tab:ddpghyp} shows individual hyperparameters of our RL agents. The hyperparameter range has been set according to common literature values and preliminary experiments on Pymgrid without ensembles. For example, our preliminary experiments teach us that less than 2 layers underperform, and more than 8 is an uncommon number of dense layers. 

\begin{table}[H]
\centering
\scriptsize
\caption{The hyperparameter spaces used for DQN agents (up) and DDPG agents (bottom)}
\addtolength{\tabcolsep}{-4pt}
\begin{tabularx}{\linewidth}{lll}
\toprule
\multicolumn{3}{c}{\normalsize DQN hyperparameters used } \\
\midrule
Name                 & Comment                                     & Value range                                                               \\
\midrule
width                & Units per layer                             & {[}32;1024{]} log2 space  \\
depth                & Number of layers                            & {[}2;8{]}                                                            \\
reward\_power        & $p$ of the reward processing                  & {[}0.5;2{]}                                                          \\
reward\_scale        & $s$ of the reward processing                  & {[}0.01;100{]} log10 \\
nb\_previous\_states & Temporal window size                        & \{1,2,4\}                                                            \\
batch\_size          & Training batch size                         & {[}16;256{]}                                                         \\
lr                   & Learning rate                               & {[}1e-4;0.1{]} log3 \\
gamma                & $\gamma$ in the bellman equation            & {[}0.2;1{]}                                                          \\
explor\_rate\_start       & First exploration rate & {[}0.06;1{]}                                                         \\
explor\_rate\_end         & Last exploration rate  & {[}0;0.05{]}                                                         \\
dueling              & Is dueling DQN enabled ?                  & \{Yes;No\}                                                           \\
\midrule
\\
\toprule
\multicolumn{3}{c}{\normalsize DDPG hyperparameters used} \\
\midrule
Name                 & Comment                                     & Value range                                                               \\
\midrule
width                & DNN width of actor and critic                            &  {[}32;1024{]} log2 \\
depth                & DNN depth of actor and critic                          & {[}2;8{]}                                                            \\
reward\_power        & $p$ of the reward processing                  & {[}0.5;2{]}                                                          \\
reward\_scale        & $s$ of the reward processing               & {[}0.01;100{]} log10 \\
nb\_previous\_states & Temporal window size                        & \{1,2,4\}                                                            \\
batch\_size          & Training batch size                         & {[}16;256{]}                                                         \\
actor\_lr                   & Learning rate of the actor                               & {[}1e-4;0.1{]} log3 \\
critic\_lr                   & Learning rate of the critic                               & {[}1e-4;0.1{]} log3 \\
gamma                & $\gamma$ in the bellman equation            & {[}0.2;1{]}                                                          \\
explor\_rate\_start       & First exploration rate & {[}0.06;1{]}                                                         \\
explor\_rate\_end         & Last exploration rate  & {[}0;0.05{]}                                                         \\
tau              &       Soft update of target parameters            & {[} 3e-4;1e-2{]}                                                          \\
\bottomrule                                                                  
\end{tabularx}
\label{tab:ddpghyp}
\end{table}

We learn from a massive amount of RL experiences the importance to process the rewards on the cumulative reward. The processed reward $r'$ is used only to train our agents, when we compare different agents we compare them based on the accumulated raw rewards.

We use this formula: $r'=sign(r)*s*abs(r)^{p}$ with $r$ the raw reward given by the environment, $s$ a scale factor and $p$ which allows smoothing ($p<1$) or increasing ($p>1$) the relative importance of the raw rewards. The $sign(r)$ function return -1 or 1 according the sign of $r$, and $abs(r)$ is the absolute value function. They allow to process reward even if the environment produces negative values.

\subsection{The ensembling procedure}

\begin{figure}[H]
  \centering
        \includegraphics[page=2,width=\linewidth,trim={0cm 11cm 21cm 0cm},clip]{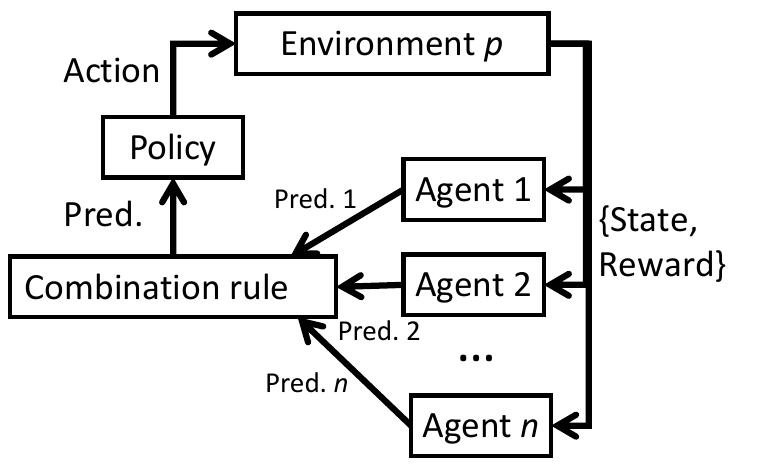}
        \caption{In the ensemble, the agents interact in parallel with the same environment. Rewards are only used to evaluate the ensemble.}
        \label{fig:ens}
\end{figure}

After training independent agents their predictions are combined to improve their ability to make better local decisions in the same environment and improve the overall trajectory as well. Figure~\ref{fig:ens} illustrates our ensemble procedure.

\subsection{The combination rule}

Multiple combination rules have been proposed by authors to combine neural network predictions. The averaging combination rule simply interpolates predictions giving the same importance to each base agent without using the validation information. The weighted averaging \cite{lin:1997} combiner uses the validation score to give more weights to the best models on the validation data but it may overfit validation information. 

The Soft Gating Principle \cite{soft:2018} is a combination method that has been proposed in supervised regression tasks to control the relative importance of the best estimators on some others. We observe it is also applicable to the reinforcement learning task with any number of base agents and both discrete and continuous action space.

The Soft Gating Principle is described by formula~\ref{eq:wj}. It computes the weight $w_{j}$ of base agent $j$  based on its validation score $R_{j}$ in the ensemble of size $J$. $\epsilon=1e-7$ is a small value above zero to avoid division by zero cases. Then, computed weights are scaled such that the sum of weights is always equal to 1. The $\beta$ value is a hyperparameter and repeated experiments will allow calibrating the combination rule. It controls the importance of better models in the final prediction. It unifies averaging ($\beta=0$), selecting only the best agent ($\beta\rightarrow+\inf $) and weighted averaging (the intermediate values for $\beta$).

\begin{equation}
w_j=\frac{ \sum\limits_{i=1}^J{R_i}  }{R_j^\beta+\epsilon} , \beta \in  \mathbb{R}_0^+
\label{eq:wj}
\end{equation}

\begin{figure}[H]
  \centering
  \includegraphics[width=\linewidth]{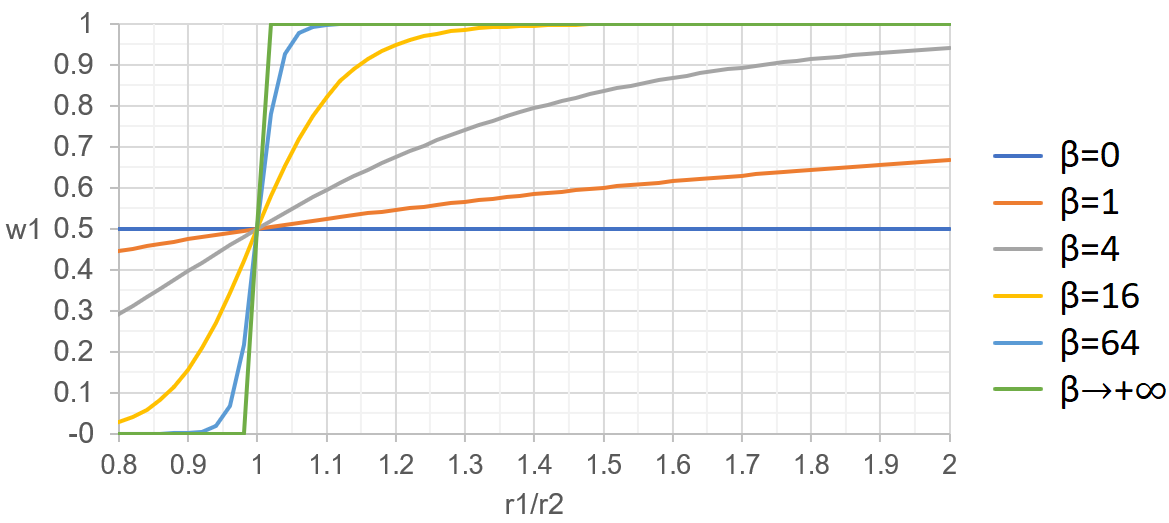}
  \caption{Agent 1's weight ($w1$) by varying to its relative score compared to agent 2 ($r1/r2$) and different value of $\beta$}
  \label{fig:bet}
\end{figure}

This equation is illustrated in figure~\ref{fig:bet}.

It suits the need to build ensemble predictions in our discrete and continuous environments, but it may not match other RL applications with continuous action space where we have physical constraints between different possible actions.

\section{Experimental settings}
\label{sec:appdesign}

In the first analysis, we arbitrarily choose the ensembles containing a fixed number of 4 base agents. We evaluate multiple ways to create ensembles: homogenous/heterogeneous and different $\beta$ values. This section details experimental settings. We use GEM \cite{rlgem} and Pymgrid \cite{rlpym} environments. Both are available with a continuous and discrete action space. We use DQN for discrete environments \cite{rlq} \cite{rldueling} and DDPG \cite{rlddpg} for continuous ones.

\textbf{GEM environment} (Gym Electric Motor) is a simulator of various electric drive motors. They contain components such as supply voltages, converters, electric motors, and load models. The goal is to maximize the performance of the engine under safety constraints. If a state exceeds the specified safety limits, the episode is stopped and the lowest possible reward is returned to the agent to punish the limit violation. In all our benchmarks we use the Permanent Magnet Synchronous Motor (``PMSMCont-v1'' and ``PMSMDisc-v1'' code name). We set the horizon to 100K steps. All our agents have been trained for 30 episodes in those environments. Validation runtime and test runtime simply run one episode but with different random seeds.

\textbf{Pymgrid environment} (PYthon MicroGRID) is a simulator of microgrids. Microgrids are contained electrical grids that are capable of disconnecting from the main grid. Automatic control allows for minimizing production costs by honoring the demand. The reward produced are negative values to maximize given by formula $r=-1*(c+p)$ with $c$ cost expressed as dollars and $p$ a penalty term when the demand for electricity experiences a power cut. In all our benchmarks we use the second microgrid available. This microgrid has multiple components connected: the demand, the solar panels, the battery, the generating set, and an unstable link with the main grid allowing to buy/sell kWh. This simulator is mixed with real sunlight time series data. The training performs 5870 steps, 438 validation steps, and 2453 test steps. Each phase receives different sunlight time series. Finally, all our agents have trained 500 episodes in those environments.

\textbf{The software stack.} We use RLLIB \cite{rllib} and  Ray \cite{ray}.

\textbf{The platform.}  We train and evaluate each population on one computing node identical to the Oak Ridge Summit node. It contains 6 Tesla-V100 GPUs and a Power9 AltiVec 2 sockets CPU, each socket containing 72 virtual CPU cores. Inference experiments have been performed on an HGX-2 containing 16 Tesla-V100 GPUs and an Intel Xeon Platinum 8168 2 sockets CPU, each one containing 48 virtual cores.

\section{Experiment results}

\subsection{Ensemble construction comparison}

We measure the performance of ensembles according to their construction (Homogeneous or Heterogeneous) and combination rules ($\beta in \{0,1,4,16,64,+\infty\}$)  . We compare them on the two simulators (Pymgrid and GEM) and the two RL algorithms (DQN and DDPG). The results are represented in table~\ref{fig:gemddpg}. 

Each strategy is evaluated on its best decile on a population of 100 ensembles with randomly drawn hyperparameters in the hyperparameter space described in table~\ref{fig:gemddpg}.


\begin{table}[H]
\small
\addtolength{\tabcolsep}{-4pt}
\caption{We compare 12 ensemble construction procedures: Homogeneous (``Homo.'') , Heterogeneous (``Hetero'') and for each one, we vary $\beta$ in \{0, 1, 4, 16, 64, +inf\}. We rank them based on their robustness to build good ensembles: the best decile of the cumulated rewards of 100 build ensembles (random hyperparameters) on the test runtime.}
\begin{tabular}{l|l|l|l|l}
Rank & DQN Pym. & DDPG Pym. & DQN GEM      & DDPG GEM    \\
\hline
 1st   & Homo. $\beta{=}0$   & Homo. $\beta{=}0$    & Homo. $\beta{=}0$    & Homo. $\beta{=}0$   \\
 2nd & Hetero. $\beta{=}4$ & Homo.  $\beta{=}4$     & Hetero. $\beta{=}0$  & Hetero. $\beta{=}0$ \\
 3rd  & Hetero. $\beta{=}1$ & Homo. $\beta{=}1$    & Hetero. $\beta{=}16$ & Homo. $\beta{=}16$
\end{tabular}

\label{fig:gemddpg}
\end{table}


\textbf{Top ensemble comparison.} When we compare the strategies based on their ability to build the best ensemble we observe no specific pattern due to random effects. This is why we use the best decile to indicate more clearly how robust a construction procedure is.

\textbf{Heterogeneous and homogeneous distribution comparison.} The first major observation is that different ensemble construction procedures yield a very different distribution of the performance.

RL agents are unstable by nature and running the same training at different runtime produces already diversified predictions. Heterogeneous ensembles combine hyperparametric and parametric diversity yielding again more diverse base agents. This diversity is maybe not useful, because it comes with the cost of much more dimensions in the hyperparameter space and is much more difficult to calibrate. In most cases, heterogeneous ensembles have a tighter distribution of $R$, for instance, the difference between the 1st and the last quartile is smaller or equal to homogeneous ensembles in all our run times.

Homogeneous is a more risky construction. If a homogeneous ensemble gets a good hyperparameter set, all four base agents have a chance to well perform but the opposite is also true: a poor hyperparameter set means all poor agents. In practice, this is not a problem that several ensembles are performing poorly, because only the ability to generate one good ensemble matters. 

Overall, a homogeneous procedure is a more robust strategy based on results in table~\ref{fig:gemddpg}. However, comparing them in more environments and more than 4 base agents should provide more insight.

\textbf{Combination rule comparison.}  Another important observation is that homogeneous ensembles with $\beta=0$ provide the best decile or equivalent decile each time compared to other $\beta$ values. This shows that the simple procedure consisting to build homogeneous ensembles and simply averaging predictions is a robust ensemble procedure.

In practice, the goal is to find the best ensemble and discard all the others. However, evaluating the procedure to build them based on only the best one is noisy information, this is why the decile is used as a measure of the robustness of the automatic ensemble construction.


%

\textbf{Distribution in discrete action space.} A last piece of information is that multiple ensembles may follow the same trajectory and accumulate the same rewards in the discrete action space.


\subsection{Stability analysis}

The previous section shows the final performance distribution with different ensemble construction procedures. It is intuitive but it is often an insufficient analysis in critical applications where stability and reproducibility are needed. Scientists want to ensure its reproducibility for transparency, confidence, and sharing experiments. After a minor update of the environment and re-training the RL algorithm, it is expected to find similar performance.

\textbf{Ensemble procedures comparison.} In table~\ref{tab:stab} the ensemble construction procedure is evaluated. First, top-3 performing hyperparameters on the validation score is selected. Then, ensembles are trained/validated/tested with 4 different random seeds. Finally, the mean and stability of the test score of the top3 ensembles are computed. More formally, they are computed with $mean(mean(R_{1,1},R_{1,2}, ...) , mean(R_{2,1},R_{2,2}, ...), ...)$ and $mean(std(R_{1,1},R_{1,2}, ...) , std(R_{2,1},R_{2,2}, ...), ..., )$ with $R_{i,j}$ is such that $i$ is the $i^{th}$ best hyperparameter (from 1 to 3), and $j$ the different run times (from 1 to 4).

\begin{table}[H]
\centering
\scriptsize
\caption{Reproducibility of the top ensembles. The blue figures indicate minimum Relative Standard Deviation value for each environment and action space.}
\addtolength{\tabcolsep}{-6pt}
\begin{tabularx}{\linewidth}{l|XX|XX}
\toprule
                & \multicolumn{2}{c|}{DQN Pymgrid}   & \multicolumn{2}{c}{DDPG Pymgrid}        \\
$\beta$                & \multicolumn{1}{c}{Homogeneous}             & \multicolumn{1}{c|}{Heterogeneous}            & \multicolumn{1}{c}{Homogeneous}             & \multicolumn{1}{c}{Heterogeneous}            \\ 
\midrule
$0$       & $-9.21M\pm35K$  & $\color{blue} -9.36M\pm7K$  & $\color{blue} -30.2M\pm0.75M$  & $-45.2M\pm5.85M$ \\

$16$      & $-9.23M\pm53K$  & $-9.14M\pm99K$ & $-33.7M\pm1.87M$ & $-40.8M\pm1.93M$ \\
$\infty$ & $-8.87M\pm183K$ & $-9.02M\pm144K$ & $-46.9M\pm4.14M$ & $-46.6M\pm5.88M$ \\
\midrule
\midrule
                & \multicolumn{2}{c|}{DQN GEM} & \multicolumn{2}{c}{DDPG GEM} \\
$\beta$                & \multicolumn{1}{c}{Homogeneous}            & \multicolumn{1}{c|}{Heterogeneous}            & \multicolumn{1}{c}{Homogeneous}           & \multicolumn{1}{c}{Heterogeneous}            \\ 
\midrule
$0$       & $\color{blue}6193\pm1066$  & $5816\pm1131$ & $3559\pm3177$  & $\color{blue}5487\pm1140$ \\
$16$      & $5837\pm1032$  & $5551\pm1147$ & $1675\pm1346$  & $3436\pm1182$ \\
$\infty$ & $5258\pm2155$  & $4496\pm2286$ & $2423\pm1775$   & $186\pm138$   \\
\bottomrule
\end{tabularx}
\label{tab:stab}
\end{table}

Comparing only the standard deviation between construction procedures is not fully relevant to observing the stability. For example, the bottom right corner experiments get a low standard deviation score compared to the others, this is because the ensemble produces a low mean score too. We propose instead the Relative Standard Deviation (RSD) given by $RSD(x)=std(x)/mean(x)$. In this case, we observe that $\beta=0$ gives generally lower RSD. It also appears that between the homogeneous and heterogeneous procedures no one seems more stable than the other in all cases.

\textbf{Ensembles compared to the best in it.}
To do it, we compare the Homogeneous ensemble with averaging as combination rule (Homogeneous and $\beta=0$) compared to the simple and common baseline consists in training 1 single RL agent (Heterogeneous and $\beta \rightarrow +\infty$). In the first strategy, 100 ensembles of 4 agents are trained, in the second one 400 agents are trained and the best is returned. 

According to table~\ref{tab:stab}, the ensembling strategy improves by a factor of 1.46 and a stability gain of 3.58.

Those two numbers are obtained by computing the median on the 4 different environments: The ensembling strategy improves by a factor of \{0.98, 1.54, 1.38, 19.1\} the cumulated rewards (e.i., the median is 1.46), and by a factor of \{4.2,5.08,2.95,0.83\} the RSD (e.i., the median is 3.58).

\begin{figure}[H]
\includegraphics[width=\linewidth]{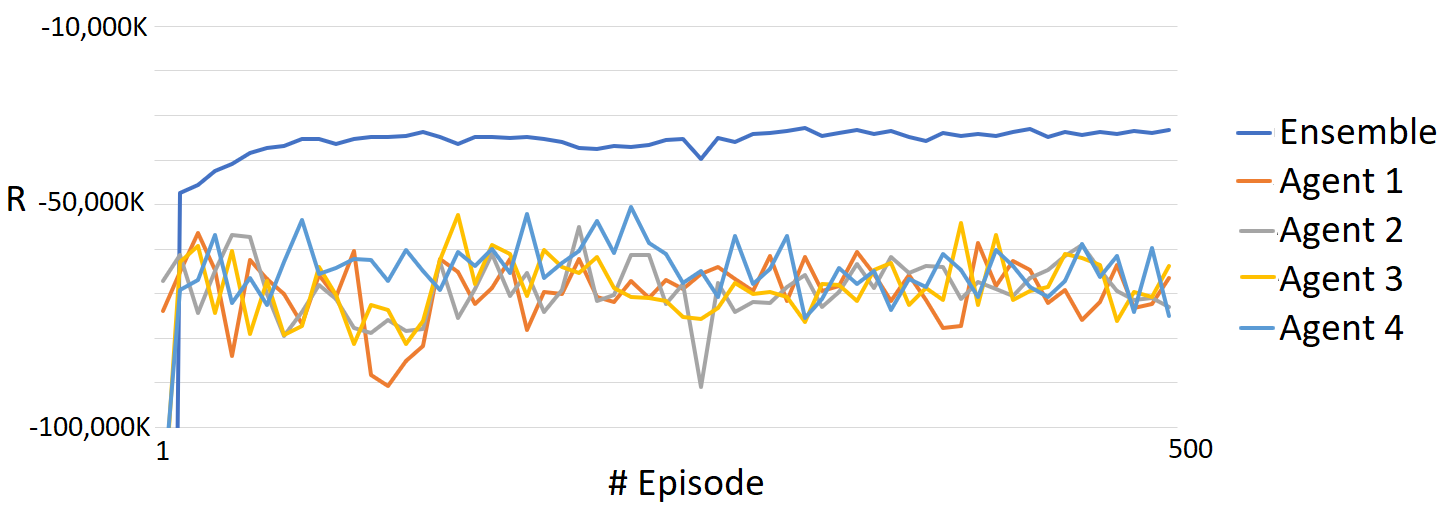}
\caption{Performance of the ensemble compared to its individual agents over 500 training episodes. After each training episode, we assess the ensemble test score and its base agents test scores. It is an example from the library of homogeneous DDPG agents on Pymgrid environment. Goal: maximization of the reward.}
\label{fig:beta}
\end{figure}

We can also observe the stability gained by ensembles compared to their base agents not only after a horizon but after each training episode as shown in figure~\ref{fig:beta}. The ensemble learning dynamics is much less noisy and may be useful to perform decisions such as early stopping \cite{early:1998} of all base agents at the same time.

\subsection{Power and time consumption at training time}

We compare different hardware allocation of agents (noted `a') and the environments (or ``simulator'' noted `s') in terms of computing time and computing power to perform the same work: we train 100 DDPG agents on Pymgrid during 1 episode, hyperparameters are drawn randomly (Random Search).

Our cluster contains 144 virtual CPU cores and 6 GPUs. Our motherboard is equipped with a sensor to measure power consumption. Some evaluated allocations are detailed as follows and results are shown in table~\ref{tab:power}: ``a/2CPUs, s/CPU'' means each agent is trained using 2 multi-threaded  CPU cores while its associated simulator is put on another CPU core.  It makes 48 agents handled at the same time. ``a/CPU, s/CPU'' means each agent is associated with one core while its associated simulator is associated with another core. It makes 72 agents handled at the same time. ``a\&s/CPU'' means each agent and its associated simulator are handled by the same core. It makes 144 agents handled at the same time. ``8a/GPU, s/CPU'' means 8 agents are co-localized in each GPU, and the corresponding simulators are associated with CPU cores. It makes 48 agents handled at the same time. Additionally, we stopped ``4a/CPU, s/CPU'' before termination because it takes too much time, and ``16a/GPU, s/CPU'' leads to memory crashes.

\begin{table}[H]
\centering
\scriptsize
\addtolength{\tabcolsep}{-5pt}
\caption{Time and energy consumed by our GPU cluster with different allocation of hardware for training 100 agents 1 episode}
\begin{tabularx}{\linewidth}{llllllll}
The assignment setting              & \begin{tabular}[c]{@{}l@{}}a/2CPU\\ s/CPU\end{tabular} & \begin{tabular}[c]{@{}l@{}}a/CPU\\ s/CPU\end{tabular} & a\&s/CPU & \begin{tabular}[c]{@{}l@{}}a/GPU\\ s/CPU\end{tabular} & \begin{tabular}[c]{@{}l@{}}2a/GPU,\\ s/CPU\end{tabular} & \begin{tabular}[c]{@{}l@{}}4a/GPU\\ s/CPU\end{tabular} & \begin{tabular}[c]{@{}l@{}}8a/GPU\\ s/CPU\end{tabular} \\
\midrule
Mean power (watt)     & 870                                                    & 875                                                   & 885      & 669                                                   & 717                                                     & 770                                                    & 1005                                                   \\
Training time (sec.)  & 928                                                    & 773                                                   & 1498     & 1791                                                  & 1217                                                    & 759                                                    & \textbf{598}                                                    \\
Energy consumed (kWh) & 0.224                                                  & 0.188                                                 & 0.368    & 0.333                                                 & 0.242                                                   & \textbf{0.162}                                                  & 0.167                                                 \\
\bottomrule
\end{tabularx}
\label{tab:power}
\end{table}

Those results show us that we can train multiple agents at the same time on a modern GPU cluster.  ``8a/GPU, s/CPU'' is the fastest allocation while ``4a/GPU, s/CPU'' is the most power-efficient. We also observe that the CPU-only implementation (``a/CPU, s/CPU'') may offer reasonable time performance without requiring GPU investment. 

\subsection{Ensemble at inference time}

Training and tuning hundreds of ensembles is a computing-intensive activity where the ensembles train batch by batch, but it is a rather time-bounded activity. The inference is a long-running activity where the ensemble is hosted in memory and waits for one single state to quickly return its associated action. In the inference phase, the agents are sequentially interacting with the environment one state and one action at a time. Therefore, the computing speed is measured with the latency defined by the elapsed time between the ensemble receiving a state from the environment, and the ensemble returning its associated action. 




The first observation is that the CPU seems more suitable for the inference phase. It is about 2 times faster than one GPU to predict the data flow. The second key observation is that increasing the number of DNNs from 1 to 8, multiplies by only 2 the prediction times. This reinforces our claim that modern computers may efficiently run these workloads. The second key observation is that increasing the number of DNNs from 1 to 8, multiplies by only 2 the prediction times. This reinforces our claim that modern computers may efficiently run these workloads. 



\section{Conclusion}
This work answers this missing piece between reinforcement learning and industrial applications aiming at stable, qualitative, and efficient algorithms.  Today, these tasks are generally solved with reinforcement learning agents without ensembles, but there is a significant improvement when multiple agents predict together. Homogeneous ensembles are less sensitive to random effects and may cumulate more rewards during the simulation. Then, we observe that modern GPUs and CPUs can efficiently train and infer an ensemble of multiple DNNs efficiently in embarrassingly parallel. Once more time in machine learning history, suitable algorithms leveraging powerful hardware allow us to solve strategic problems.



\section{Future works and distribution shift}



Electric simulators may be developed with a high degree of realism but some unexpected and rare events may occur. A key machine learning skill in dealing with such situations is that the model recognizes its ignorance when unexpected states occur, as well as complicated states with a high risk to compute a poor action. In addition to the performance and stability gain, ensembles are known  \cite{uncertainty:2017} \cite{fort2021} \cite{rlcio:2015} to produce well-calibrated uncertainty estimates. Further analysis should investigate how well uncertainty estimate is reliable in the context of RL.



\bibliography{bib_automl, bib_ensemble, bib_hpml, bib_infer, bib_rl, bib_uncertainty}

\begin{thebibliography}{10}
\providecommand{\url}[1]{#1}
\csname url@samestyle\endcsname
\providecommand{\newblock}{\relax}
\providecommand{\bibinfo}[2]{#2}
\providecommand{\BIBentrySTDinterwordspacing}{\spaceskip=0pt\relax}
\providecommand{\BIBentryALTinterwordstretchfactor}{4}
\providecommand{\BIBentryALTinterwordspacing}{\spaceskip=\fontdimen2\font plus
\BIBentryALTinterwordstretchfactor\fontdimen3\font minus
  \fontdimen4\font\relax}
\providecommand{\BIBforeignlanguage}[2]{{%
\expandafter\ifx\csname l@#1\endcsname\relax
\typeout{** WARNING: IEEEtran.bst: No hyphenation pattern has been}%
\typeout{** loaded for the language `#1'. Using the pattern for}%
\typeout{** the default language instead.}%
\else
\language=\csname l@#1\endcsname
\fi
#2}}
\providecommand{\BIBdecl}{\relax}
\BIBdecl

\bibitem{rlpym}
\BIBentryALTinterwordspacing
G.~Henri, T.~Levent, A.~Halev, R.~Alami, and P.~Cordier, ``pymgrid: An
  open-source python microgrid simulator for applied artificial intelligence
  research,'' in \emph{NeurIPS 2020 Workshop on Tackling Climate Change with
  Machine Learning}, 2020. [Online]. Available:
  \url{https://www.climatechange.ai/papers/neurips2020/3.html}
\BIBentrySTDinterwordspacing

\bibitem{rlanm:2021}
\BIBentryALTinterwordspacing
R.~Henry and D.~Ernst, ``Gym-anm: Open-source software to leverage
  reinforcement learning for power system management in research and
  education,'' \emph{Software Impacts}, vol.~9, p. 100092, 2021. [Online].
  Available:
  \url{https://www.sciencedirect.com/science/article/pii/S2665963821000348}
\BIBentrySTDinterwordspacing

\bibitem{rlgem}
A.~{Traue}, G.~{Book}, W.~{Kirchgässner}, and O.~{Wallscheid}, ``Toward a
  reinforcement learning environment toolbox for intelligent electric motor
  control,'' \emph{IEEE Transactions on Neural Networks and Learning Systems},
  pp. 1--10, 2020.

\bibitem{mnih:2015}
\BIBentryALTinterwordspacing
V.~Mnih, K.~Kavukcuoglu, D.~Silver, A.~A. Rusu, J.~Veness, M.~G. Bellemare,
  A.~Graves, M.~Riedmiller, A.~K. Fidjeland, G.~Ostrovski, S.~Petersen,
  C.~Beattie, A.~Sadik, I.~Antonoglou, H.~King, D.~Kumaran, D.~Wierstra,
  S.~Legg, and D.~Hassabis, ``Human-level control through deep reinforcement
  learning,'' \emph{Nature}, vol. 518, no. 7540, pp. 529--533, Feb. 2015.
  [Online]. Available: \url{http://dx.doi.org/10.1038/nature14236}
\BIBentrySTDinterwordspacing

\bibitem{silver:2016}
\BIBentryALTinterwordspacing
D.~Silver, A.~Huang, C.~J. Maddison, A.~Guez, L.~Sifre, G.~van~den Driessche,
  J.~Schrittwieser, I.~Antonoglou, V.~Panneershelvam, M.~Lanctot, S.~Dieleman,
  D.~Grewe, J.~Nham, N.~Kalchbrenner, I.~Sutskever, T.~Lillicrap, M.~Leach,
  K.~Kavukcuoglu, T.~Graepel, and D.~Hassabis, ``Mastering the game of go with
  deep neural networks and tree search,'' \emph{Nature}, vol. 529, no. 7587,
  pp. 484--489, Jan 2016. [Online]. Available:
  \url{https://doi.org/10.1038/nature16961}
\BIBentrySTDinterwordspacing

\bibitem{rlapplications:2021}
\BIBentryALTinterwordspacing
A.~Perera and P.~Kamalaruban, ``Applications of reinforcement learning in
  energy systems,'' \emph{Renewable and Sustainable Energy Reviews}, vol. 137,
  p. 110618, 2021. [Online]. Available:
  \url{https://www.sciencedirect.com/science/article/pii/S1364032120309023}
\BIBentrySTDinterwordspacing

\bibitem{rlmatter}
P.~Henderson, R.~Islam, P.~Bachman, J.~Pineau, D.~Precup, and D.~Meger, ``Deep
  reinforcement learning that matters,'' in \emph{AAAI}, 2018.

\bibitem{useoverfit:1995}
P.~Sollich and A.~Krogh, ``Learning with ensembles: How overfitting can be
  useful,'' in \emph{NIPS}, 1995.

\bibitem{ensrldqn}
D.~L. Elliott and C.~Anderson, ``The wisdom of the crowd: Reliable deep
  reinforcement learning through ensembles of q-functions,'' \emph{IEEE
  Transactions on Neural Networks and Learning Systems}, pp. 1--9, 2021.

\bibitem{rlmetrpo:2018}
\BIBentryALTinterwordspacing
T.~Kurutach, I.~Clavera, Y.~Duan, A.~Tamar, and P.~Abbeel, ``Model-ensemble
  trust-region policy optimization,'' in \emph{International Conference on
  Learning Representations}, 2018. [Online]. Available:
  \url{https://openreview.net/forum?id=SJJinbWRZ}
\BIBentrySTDinterwordspacing

\bibitem{heteroinput:1999}
Y.~Liao and J.~Moody, ``Constructing heterogeneous committees using input
  feature grouping: Application to economic forecasting,'' p. 921–927, 1999.

\bibitem{creatediversity:2005}
G.~Brown, J.~Wyatt, R.~Harris, and X.~Yao, ``Diversity creation methods: A
  survey and categorisation,'' \emph{Information Fusion}, vol.~6, pp. 5--20, 03
  2005.

\bibitem{wavg:1994}
S.~Hashem, B.~Schmeiser, and Y.~Yih, ``Optimal linear combinations of neural
  networks: an overview,'' in \emph{Proceedings of 1994 IEEE International
  Conference on Neural Networks (ICNN'94)}, vol.~3, 1994, pp. 1507--1512 vol.3.

\bibitem{rlq}
H.~v. Hasselt, A.~Guez, and D.~Silver, ``Deep reinforcement learning with
  double q-learning,'' in \emph{Proceedings of the Thirtieth AAAI Conference on
  Artificial Intelligence}, ser. AAAI'16.\hskip 1em plus 0.5em minus
  0.4em\relax AAAI Press, 2016, p. 2094–2100.

\bibitem{rldueling}
Z.~Wang, T.~Schaul, M.~Hessel, H.~Van~Hasselt, M.~Lanctot, and N.~De~Freitas,
  ``Dueling network architectures for deep reinforcement learning,'' in
  \emph{Proceedings of the 33rd International Conference on International
  Conference on Machine Learning - Volume 48}, ser. ICML'16.\hskip 1em plus
  0.5em minus 0.4em\relax JMLR.org, 2016, p. 1995–2003.

\bibitem{rlddpg}
\BIBentryALTinterwordspacing
T.~Xu, Q.~Liu, L.~Zhao, and J.~Peng, ``Learning to explore via meta-policy
  gradient,'' in \emph{Proceedings of the 35th International Conference on
  Machine Learning}, ser. Proceedings of Machine Learning Research, J.~Dy and
  A.~Krause, Eds., vol.~80.\hskip 1em plus 0.5em minus 0.4em\relax PMLR, 10--15
  Jul 2018, pp. 5463--5472. [Online]. Available:
  \url{https://proceedings.mlr.press/v80/xu18d.html}
\BIBentrySTDinterwordspacing

\bibitem{random:2020}
\BIBentryALTinterwordspacing
L.~Li and A.~Talwalkar, ``Random search and reproducibility for neural
  architecture search,'' in \emph{Proceedings of The 35th Uncertainty in
  Artificial Intelligence Conference}, ser. Proceedings of Machine Learning
  Research, R.~P. Adams and V.~Gogate, Eds., vol. 115.\hskip 1em plus 0.5em
  minus 0.4em\relax PMLR, 22--25 Jul 2020, pp. 367--377. [Online]. Available:
  \url{http://proceedings.mlr.press/v115/li20c.html}
\BIBentrySTDinterwordspacing

\bibitem{early:1998}
L.~Prechelt, ``Early stopping-but when?'' in \emph{Neural Networks: Tricks of
  the Trade, This Book is an Outgrowth of a 1996 NIPS Workshop}.\hskip 1em plus
  0.5em minus 0.4em\relax Berlin, Heidelberg: Springer-Verlag, 1998, p.
  55–69.

\bibitem{lin:1997}
\BIBentryALTinterwordspacing
S.~Hashem, ``Optimal linear combinations of neural networks,'' \emph{Neural
  Netw.}, vol.~10, no.~4, p. 599–614, Jun. 1997. [Online]. Available:
  \url{https://doi.org/10.1016/S0893-6080(96)00098-6}
\BIBentrySTDinterwordspacing

\bibitem{soft:2018}
\BIBentryALTinterwordspacing
A.~Arpteg, B.~Brinne, L.~Crnkovic{-}Friis, and J.~Bosch, ``Software engineering
  challenges of deep learning,'' \emph{CoRR}, vol. abs/1810.12034, 2018.
  [Online]. Available: \url{http://arxiv.org/abs/1810.12034}
\BIBentrySTDinterwordspacing

\bibitem{rllib}
E.~Liang, ``Scalable reinforcement learning systems and their applications,''
  EECS, university of California, Berkley, Tech. Rep., 2021.

\bibitem{ray}
P.~Moritz, R.~Nishihara, S.~Wang, A.~Tumanov, R.~Liaw, E.~Liang, M.~Elibol,
  Z.~Yang, W.~Paul, M.~I. Jordan, and I.~Stoica, ``Ray: A distributed framework
  for emerging ai applications,'' in \emph{Proceedings of the 13th USENIX
  Conference on Operating Systems Design and Implementation}, ser.
  OSDI'18.\hskip 1em plus 0.5em minus 0.4em\relax USA: USENIX Association,
  2018, p. 561–577.

\bibitem{uncertainty:2017}
B.~Lakshminarayanan, A.~Pritzel, and C.~Blundell, ``Simple and scalable
  predictive uncertainty estimation using deep ensembles,'' in \emph{NIPS},
  2017.

\bibitem{fort2021}
\BIBentryALTinterwordspacing
M.~Havasi, R.~Jenatton, S.~Fort, J.~Z. Liu, J.~Snoek, B.~Lakshminarayanan,
  A.~M. Dai, and D.~Tran, ``Training independent subnetworks for robust
  prediction,'' in \emph{9th International Conference on Learning
  Representations, {ICLR} 2021, Virtual Event, Austria, May 3-7, 2021}.\hskip
  1em plus 0.5em minus 0.4em\relax OpenReview.net, 2021. [Online]. Available:
  \url{https://openreview.net/forum?id=OGg9XnKxFAH}
\BIBentrySTDinterwordspacing

\bibitem{rlcio:2015}
I.~Mordatch, K.~Lowrey, and E.~Todorov, ``Ensemble-cio: Full-body dynamic
  motion planning that transfers to physical humanoids,'' in \emph{2015
  IEEE/RSJ International Conference on Intelligent Robots and Systems (IROS)},
  2015, pp. 5307--5314.

\end{thebibliography}
\bibliographystyle{IEEEtran}




\end{document}